\documentclass{article}

\usepackage[nonatbib,final]{neurips_2025}

\usepackage[utf8]{inputenc} 
\usepackage[T1]{fontenc}    
\usepackage{hyperref}       
\usepackage{url}            
\usepackage{booktabs}       
\usepackage{amsfonts}       
\usepackage{nicefrac}       
\usepackage{microtype}      
\usepackage{xcolor}         

\usepackage{arydshln}
\usepackage{times}
\usepackage{soul}
\usepackage[small]{caption}
\usepackage{graphicx}
\usepackage{amsmath}
\usepackage{amsthm}
\usepackage{algorithm}
\usepackage{algorithmic}
\usepackage{lineno}
\usepackage{multirow} 
\usepackage{amssymb}
\usepackage{color}

\title{iTFKAN: Interpretable Time Series Forecasting with Kolmogorov–Arnold Network}

\author{%
  Ziran Liang\\
  Hong Kong Polytechnic University\\
  \texttt{ziran.liang@connect.polyu.hk} \\
  \And
  Rui An \\
  Hong Kong Polytechnic University \& Northwestern Polytechnical University\\
  \texttt{rui77.an@connect.polyu.hk} \\
  \And
  Wenqi Fan\thanks{Corresponding Author\\This work has been submitted to the lEEE for possible publication. Copyright may be transferred without notice, after which this version may no longer be accessible.} \\
  Hong Kong Polytechnic University\\
  \texttt{wenqifan03@gmail.com} \\
  \And
  Yanghui Rao \\
  Sun Yat-sen University\\
  \texttt{raoyangh@mail.sysu.edu.cn} \\
  \And
  Yuxuan Liang \\
  Hong Kong University of Science and Technology (Guangzhou)\\
  \texttt{yuxliang@outlook.com} \\
}

\begin{document}

\maketitle

\begin{abstract}

As time evolves, data within specific domains exhibit predictability that motivates time series forecasting to predict future trends from historical data.
However, current deep forecasting methods can achieve promising performance but generally lack interpretability, hindering trustworthiness and practical deployment in safety-critical applications such as auto-driving and healthcare.
In this paper, we propose a novel interpretable model, \textbf{iTFKAN}, for credible time series forecasting.
iTFKAN enables further exploration of model decision rationales and underlying data patterns due to its interpretability achieved through model symbolization.
Besides, iTFKAN develops two strategies, prior knowledge injection, and time-frequency synergy learning, to effectively guide model learning under complex intertwined time series data. 
Extensive experimental results demonstrated that iTFKAN can achieve promising forecasting performance while simultaneously possessing high interpretive capabilities.

\end{abstract}

 \section{Introduction}

Time series forecasting is a vital task in AI~\cite{ning2025surveywebagentsnextgenerationai,liu2022trustworthy}, based on the assumption that time series data exhibits predictability,  involving the mining of latent temporal patterns and the prediction of future trends based on past historical data. 
In practical scenarios, time series data in the forecasting task often contains complex temporal patterns, including seasonality, trends, random fluctuation, and sudden changes.
These patterns often intertwine with each other,  leading to complex and dynamic behaviors, particularly in various real-world applications such as financial market analysis~\cite{financialtsf}, energy demand forecasting~\cite{energytsf,li2025urbanev}, weather prediction~\cite{weathertsf}, and traffic flow forecasting~\cite{traffictsf}.
Recently, the development in deep learning has catalyzed the numerous creations of deep time series forecasting models, including those based on Recurrent Neural Network (RNN)~\cite{thoc,timegrad}, Convolutional Neural Network (CNN)~\cite{scinet,micn,timesnet}, Multi-Layer Perceptron (MLP)~\cite{dlinear,tide,timemixer}, and Transformers~\cite{itransformer,autoformer,etsformer}.
These deep forecasters, especially MLP-based and Transformer-based ones, achieved superior performance compared to traditional statistical forecasters.
However, the forecasting processes of these methods often lack interpretability due to their highly complex network structures.
Besides, there is no explicit symbolic representation (e.g., rules, formulas, or tree structures) of the decision-making process.
Therefore, the forecasting processes of these methods often lack interpretability and transparency, which makes it difficult to track model decision rationales.
The absence of interpretability makes it difficult for researchers and practitioners to understand the underlying rationales behind the model's predictions, posing huge risks to their usage in critical applications like healthcare~\cite{health_interpre} and finance ~\cite{finance_interpre,finance_interpre1} where the consequences of incorrect decisions can cause severe economic and security consequences. Therefore, there is an urgent need to develop models that not only deliver high prediction performance but also exhibit robust interpretability.

As one of the most advanced AI techniques, Kolmogorov–Arnold Network (\textbf{KAN})~\cite{kan,kan2} emerges as a promising alternative to the MLP, opening up a new possibility for AI models' interpretability~\cite{explainable_survey, ning2024joint,fan2023jointly}.
Technically, KAN replaces the traditional fixed activation functions with learnable activation functions, i.e., univariate spline parameterized functions, and performs summation operations on the nodes.
Therefore, KAN allows the behavior of each node to be explained intuitively in the form of its activation functions and summation operations.
This interpretability is not limited to the node level, but can be extended to the entire network structure since its theoretical foundations guarantee that the network can be approximated by a finite number of univariate functions and summing operations. 
In that way, people can analyze these univariate activation functions to understand the working mechanics of the entire model, which greatly enhances the transparency and trustworthiness of the model.
Its interpretability has gained widespread application, providing a clear understanding path for tasks such as dynamical systems~\cite{kanode}, computational biomedicine~\cite{biokan}, and tabular operation~\cite{tablekan}.
Therefore, given their great advantages, KAN provides great opportunities to advance the interpretability of time series forecasting. 

Despite its potential for credible time series analysis, the application of KAN in time series forecasting still faces some challenges.
The complex intertwined patterns inherent in time series might hinder the ability of KAN to spontaneously learn ideal model structures and consequently affect its prediction performance, thus requiring additional guidance~\cite{karsein,cfkan}.
Accordingly, to better adapt KAN's powerful interpretive capabilities to the time series domain, we enhance KAN by 1) incorporating prior knowledge and 2) harnessing the complementary advantages of time and frequency perspectives.
For the former, given KAN's ability to bridge symbolism and connectionism, we propose injecting prior knowledge to orient its model structure learning, where the prior knowledge takes the form of symbolic formulas whose characteristics match some patterns of time series. 
For the latter, we develop a time-frequency synergy learning strategy to disentangle complex patterns in time series from multiple perspectives, in which the time perspective provides local sequential information of time points, while the frequency perspective offers global periodic information. Note that the synergy learning strategy of two different perspectives can effectively prevent information loss as well as combine their strengths.

In this paper, we propose a \textbf{i}nterpretable method 
built on \textbf{KAN} that incorporates symbolic injection and \textbf{T}ime-\textbf{F}requency synergy learning, \textbf{iTFKAN}. 
iTFKAN is endowed with interpretability owing to the model design of KAN, which facilitates people to understand the decision rationales and find underlying data patterns of time series for trustworthy time series forecasting.
In addition, prior knowledge injection empowers iTFKAN through characterized symbolic formulas, which efficiently guide the model by assimilating prior knowledge.
Furthermore, time-frequency synergy learning disentangles the complicated patterns from both time and frequency perspectives, thus facilitating iTFKAN to fully utilize their complementary advantages.
The main contribution can be summarized as follows:
\begin{itemize}
\item We propose a novel interpretable framework, \textbf{iTFKAN}, for time series forecasting, which harnesses the power of KAN to effectively model time series data with symbolization and reflect the prediction rationales inside the model and latent patterns in data.

\item We propose two principled strategies, prior knowledge injection, and time-frequency synergy learning, to sufficiently guide model learning and achieve credible and effective time series forecasting. 
\item Our proposed iTFKAN exhibits robust interpretability by providing prediction rationals and achieves superior performance over various benchmarks, which establishes solid a foundation for its further application in real-world scenarios.
\end{itemize}

\section{Related Work}
\label{sec:relatedwork}

\paragraph{Deep Time Series Forecasting models.} 
The emergence of deep learning has marked a pivotal shift in time series forecasting, transitioning from traditional statistical forecasters to forecasters based on deep learning methods. 
These include CNN-based models \cite{timesnet}, Transformer-based models \cite{autoformer,informer,itransformer}, and MLP-based models\cite{dlinear,timemixer}.
Among these, CNN-based methods such as TimesNet~\cite{timesnet} and SCINet~\cite{scinet} benefit from the strong capabilities of convolution in extracting local features. 
These models can effectively learn local non-stationary patterns and multi-level information through carefully designed hierarchical architectures. Transformer-based methods, like PatchTST~\cite{patchtst} and iTransformer~\cite{itransformer}, capitalize on the advantages of the self-attention mechanism to capture long-term dependencies, demonstrating strong performance in time series forecasting. These methods have become the dominant architectures for time series analysis.
Recently, MLP-based methods such as Dlinear~\cite{dlinear} and TimeMixer~\cite{timemixer} have gained attention due to the point-wise mapping characteristic of linear layers, which makes them adept at learning time-step dependent features. Moreover, their efficiency and simplicity have challenged the long-dominant Transformer-based forecasters, triggering a boom in MLP-based models. Unfortunately, neither of these deep forecasters is interpretable, which inhibits the credibility of the prediction process.

\paragraph{Kolmogorov–Arnold Network (KAN) for Interpretability.}
Recently, KAN has garnered considerable attention for its interpretability in its predictions.
KAN has been successfully applied to various fields where interpretability is crucial, such as medical image processing~\cite{ukan}, Internet of Things (IoT)~\cite{iotkan}, physics~\cite{ginnkan}, concept drift~\cite{kan4drift}, etc. 
Some efforts have been made to directly apply KAN for time series forecasting~\cite{simplekan2,timekan2},  without further analyzing or exploring the internal prediction mechanisms and key details of KAN.
Han et al.~\cite{kan4tsf} employ gating mechanisms to provide interpretability but completely ignores KAN's inherent interpretability. 
Some existing works focus on the interpretability of the relationships between variables~\cite{solarkan,tsclassifi}.
Therefore, while there have been some preliminary investigations into the use of KAN for time series forecasting tasks~\cite{simplekan1,simplekan3,bilstmkan}, its potential for interpretability remains largely underexplored.
Differentiating from these models, we aim to leverage the power of KAN to uncover both the decision-making rationales of the model and the underlying data patterns in time series forecasting — a critical yet underexplored challenge. Our proposed iTFKAN provides interpretable solutions for time series forecasting, with significant practical value.

\section{Preliminaries: Kolmogorov–Arnold Network (KAN)}

\label{section:kan}
KAN emerges as a promising alternative to the MLP, boasting both accuracy and interpretability in function fitting and PDE solving area~\cite{kan}. 
The theoretical foundation of KAN stems from the \textit{Kolmogorov-Arnold Representation Theorem}, which established that a continuous function $f$ on a bounded domain, with multiple variates $\mathbf{x}=(x_1,\cdots,x_n)$, can be written as a finite composition of continuous functions of a single variable and the binary operation of addition:  
\begin{equation}
\begin{aligned}
f(\mathbf{x})=f(x_1,\cdots,x_n)=\sum_{q=1}^{2n+1}\Phi_q\left(\sum_{p=1}^n\phi_{q,p}(x_p)\right),
\end{aligned}
\end{equation}
where $\phi_{q,p}$ and $\Phi_q$ are univariate functions at different hierarchies.
Based on the Kolmogorov-Arnold Representation Theorem, the architecture of KAN is designed as follows:
\begin{equation}
\begin{split}
&\mathrm{KAN}(\mathbf{x}) =(\mathbf{\Phi}^D \circ \cdots\circ \mathbf{\Phi}^2 \circ \mathbf{\Phi}^1)(\mathbf{x}),\\
&\mathbf{\Phi}^d = \{\phi^d_{i,j} \mid 1 \leq i \leq I, 1 \leq j \leq J\},
\end{split}
\end{equation}
where $D$ denotes the depth of the KAN network.  
$\mathbf{\Phi}^d $ represents all total univariate activation functions for the $d$-th layer in KAN.  $I$ and $J$ denote the input and output dimensions of the KAN at that layer, respectively. Generally, each univariate activation function is parametrized as a \emph{B-spline curve} in  KAN.
Different from the architecture of general MLP approaches (i.e., training learnable weights on edges and employing fixed activation functions on nodes), KAN innovatively trains learnable univariate activation functions directly on edges. 
The paradigm shift and theoretical foundation collectively endow KAN with remarkable \textbf{interpretability}~\cite{kan2}.

\begin{figure*}[t!]
	\centering
	\includegraphics[width=\linewidth]{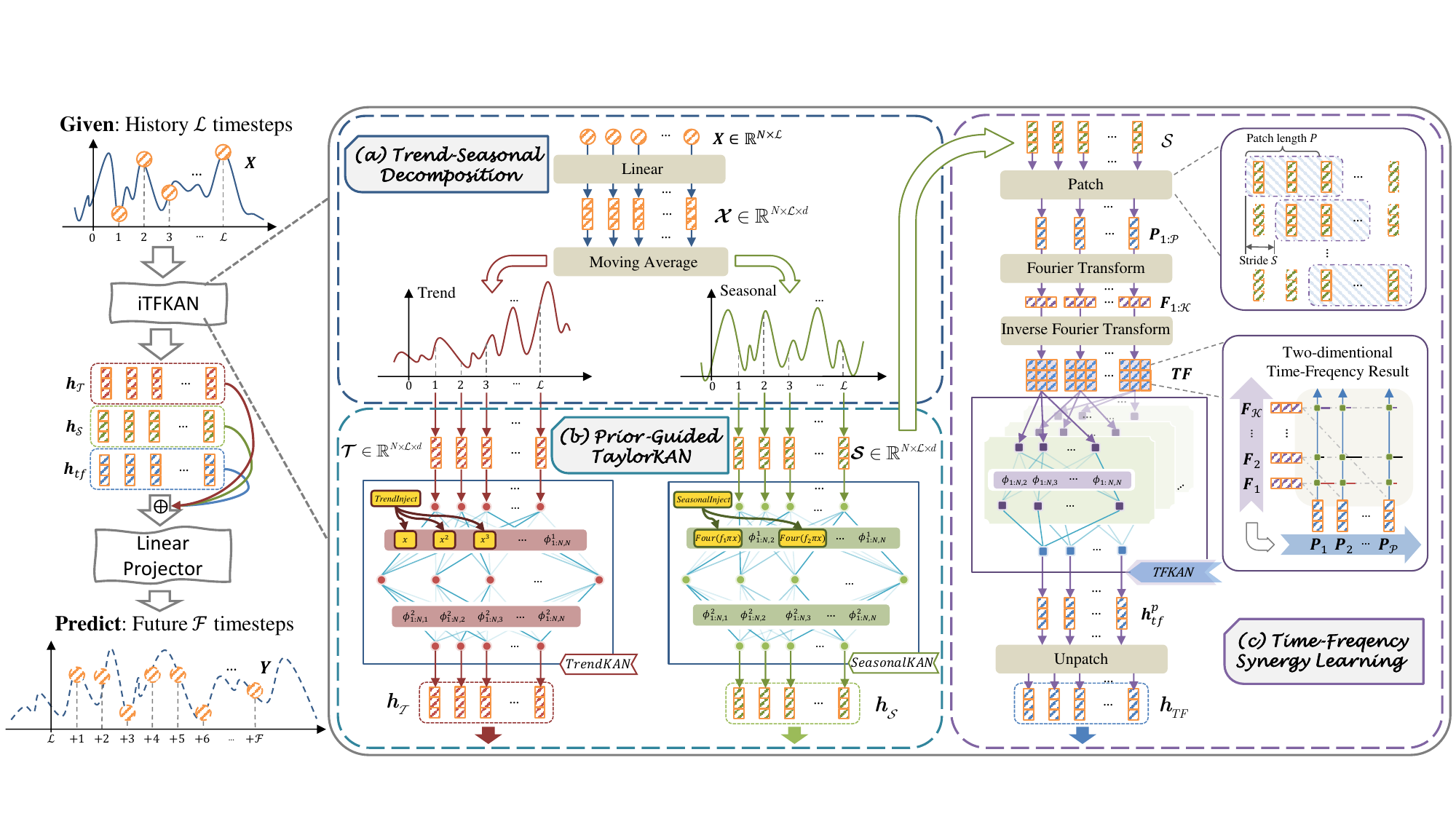}
	\caption{The overall framework of iTFKAN. The iTFKAN comprises three key components: \textit{Trend-Seasonal Decomposition}, \textit{Prior-Guided TaylorKAN}, and \textit{Time-Frequency Synergy Learning}. Each module is specifically designed to learn distinct patterns: trend, seasonal, and time-frequency dependencies, respectively. Finally, these three informative representations are linearly mapped to output the forecast results.
    }
	\label{fig:architecture}
\end{figure*}

\section{Methodology}
\label{sec:methodlogy}

\subsection{Overview}

Given a series time points $\mathbf{X}=\{\mathbf{x}_1,\mathbf{x}_2,\cdots,\mathbf{x}_t,\cdots,\mathbf{x}_{\mathcal{L}}\}\in\mathbb{R}^{N\times \mathcal{L}}$, where $\mathbf{x}_t\in\mathbb{R}^{N}$ represents $N$ distinct series at timestamp $t$. The goal of time series forecasting is to predict the future $\mathcal{F}$ timestamps $\mathbf{Y}=\{\mathbf{x}_{\mathcal{L}+1},\mathbf{x}_{\mathcal{L}+2},\cdots,\mathbf{x}_{\mathcal{L}+\mathcal{F}}\}\in\mathbb{R}^{N\times \mathcal{F}}$ based on historical observations $\mathbf{X}$. 
Due to the complex and intertwined patterns observed in time series, the adoption of KAN presents significant challenges.
To address this issue, we propose a novel framework, iTFKAN, for time series forecasting. As illustrated in Figure~\ref{fig:architecture}, the proposed model consists of three key modules: \emph{Trend-Seasonal Decomposition}, \emph{Prior-Guided TaylorKAN}, and \emph{Time-Frequency Synergy Learning}. The Trend-Seasonal Decomposition module decomposes the raw time series into trend and seasonal components, facilitating their subsequent specialized processing. Then, the Prior-Guided TaylorKAN module incorporates prior knowledge through symbolic formulas to guide the structured learning process.
Finally, the Time-Frequency Synergy Learning module systematically performs multi-perspective learning from both time and frequency domains, providing a comprehensive view to obtain informative representations.
After iTFKAN, a linear projector is utilized to map the learned representations for generating the final predictions for various applications.

\newcommand{\Ein}{{\vE_{\text{in}}}}

\subsection{Trend-Seasonal Decomposition}

\label{sec:basic_architecture}
Time series data often exhibit complex, intertwined patterns that can be broadly categorized into trend and seasonal~\cite{stl}. The trend pattern reflects long-term dynamics such as general increases, decreases, or stability over time, while the seasonal pattern represents recurring cycles at regular intervals, such as daily, monthly, or yearly. To effectively model these patterns, a feasible approach involves decomposing the time series into trend and seasonal components and modeling them separately. This decomposition simplifies the overall learning task by reducing complexity and providing more targeted inputs for downstream modeling processes~\cite{cao2024tempo}.
Building upon this insight, we first map the input $\mathbf{X}\in\mathbb{R}^{N\times \mathcal{L}}$ to a high-dimensional and more expressive representation $\mathcal{X} \in\mathbb{R}^{N\times \mathcal{L}\times d}$ through a linear layer.
Then, we adapt the Moving Average strategy~\cite{autoformer} to decompose the high-dimensional representation $\mathcal{X}$ into two meaningful component:
\begin{equation}
\begin{split}
\text{Trend Pattern: } &\mathcal{T}=\text{AvgPool(Padding}(\mathcal{X})),\\
\text{Seasonality Pattern: } &\mathcal{S}=\mathcal{X} - \mathcal{T}.
\end{split}
\end{equation}
where $\mathcal{T}$ and $ \mathcal{S}\in\mathbb{R}^{N\times \mathcal{L}\times d}$ denote the trend and seasonal components, respectively. The AvgPool($\cdot$) aims to smooth out short-term fluctuations and emphasizes long-term trends.

\subsection{Prior-Guided TaylorKAN}
To efficiently extract distinctive features from the trend component $\mathcal{T}$ and the seasonal component $\mathcal{S}$, we propose the Prior-Guided TaylorKAN to enhance the original KAN from computational efficiency and model structure optimization, which can be achieved by Taylor Series Expansion and Prior Knowledge Injection, respectively.
\paragraph{Taylor Series Expansion.}
In general, the typical KAN employs a spline-parameterized univariate activation function, which incurs high computational complexity due to its iterative calculation. 
To alleviate this problem, 
we take advantage of the Taylor series expansion to precisely model the univariate activation function $\phi(x)$ on each edge with a low computational burden. 
Thus, the activation function in KAN can be formulated as:
\begin{equation}
    \label{equa:taylor4}
    \phi(x)=w(b(x)+\text{Taylor}(x))=w(b(x)+\sum_{o=0}^Oa_ox^o),
\end{equation}
where $b(\cdot)$ is a basis function, $w$ is a trainable weight, and $a_o$ denotes the coefficient of $x^o$.
As second-order Taylor expansion functions already excel in local approximation, the highest order of the Taylor series expansion is set to 2 here.

\paragraph{Prior Knowledge Injection.}
\label{sec:symbolic_injection}
Meanwhile, the general KAN faces challenges in spontaneously learning the ideal model structure, as it is prone to get trapped in local optima when dealing with complex patterns. Therefore, injecting symbolic formulas is a practical strategy to alleviate the dependence of KAN on data-driven optimization and to utilize prior knowledge to guide the learning direction~\cite{karsein,cfkan}. Inspired by these insights, we propose formalizing the prior knowledge in time series—specifically the long-term monotonicity of the trend component and the repeating cycles of the seasonal component—as symbolic formulas, which are then injected into the activation functions of KAN.
Specifically, \textit{TrendInject} and \textit{SeasonalInject} strategies are introduced to utilize a polynomial of small degree $p$ 
for modeling the monotonicity of the trend component, and employ a Fourier series for modeling the periodicity of the seasonal component, as expressed in:
\begin{equation}
\label{equa:mono}
\textit{TrendInject}(x)=m_px^p+m_{p-1}x^{p-1}+\cdots+m_1x+m_0,
\end{equation}
\begin{equation}
\label{equa:period}
\textit{SeasonalInject}(x)=\frac{a_0}{2}+\sum_{k=1}^K\left[a_k\mathrm{cos}(f_k\pi x)+b_k\mathrm{sin}(f_k\pi x)\right],
\end{equation}
where $\mathbf{m} = [m_0, m_1, \cdots, m_{p-1}, m_p]$ represents the polynomial coefficients corresponding to the respective powers of \(x\). $f_k$ denotes one of the top-$K$ frequencies extracted by Fourier Transform, highlighting the most significant periodic patterns.
$\mathbf{a}=[a_0, a_1, \cdots, a_{K}]$ and $\mathbf{b}=[b_1, \cdots, b_{K}]$ denote the corresponding trainable coefficients of frequency $f_k$.

The \textit{TrendInject} and \textit{SeasonalInject}
serve as the sum of univariate symbolic formulas. We then set these symbolic formulas to the corresponding activation functions and propose \textit{TrendKAN} and \textit{SeasonalKAN}, where the activation can be reformulated as:
\begin{equation}
\mathbf{\Phi}^1_{Trend} = \{x^j \mid 1 \leq i \leq I, 1 \leq j \leq p\},
\end{equation}
\begin{equation}
\begin{aligned}
 \mathbf{\Phi}_{Season}^1=\{\mathrm{Four}(f_j\pi x)\mid1&\leq i\leq I,j\in\mathcal{I}_K\},
\end{aligned}
\end{equation}
where $\mathrm{Four}(f_k\pi x)=\mathrm{cos}(f_k\pi x)+\mathrm{sin}(f_k\pi x)$, $\mathcal{I}_K=\{i_1,i_2,\cdots,i_K\}$ denotes the indices of the top-$K$ frequencies. It is worth noting that all symbolic formula settings are located at the first layer, allowing the KANs to further adapt them in the next layer.

Finally, the trend component $\mathcal{T}$ and the seasonal component $\mathcal{S}$ are fed into the \textit{TrendKAN} and \textit{SeasonalKAN} to learn non-linear representations of time series information as follows:
\begin{equation}
     \mathbf{h}_{\mathcal{T}}=\textit{TrendKAN}(\mathcal{T}),
     \mathbf{h}_{\mathcal{S}}=\textit{SeasonalKAN}(\mathcal{S}),
 \end{equation}
where $\mathbf{h}_{\mathcal{T}}$ and $\mathbf{h}_{\mathcal{S}}\in\mathbb{R}^{N\times \mathcal{L}\times d}$ denote the distinctive representations learned from trend and seasonal component, respectively.

\subsection{Time-Frequency Synergy Learning}
\label{sec:tf}

It is well-recognized that the time domain and frequency domain describe time series information from two distinct perspectives~\cite{frets}. 
The time domain focuses on sequential relationships between data points, providing insights into temporal patterns, while the frequency domain offers a global perspective, highlighting periodicity. 
Efforts have been made to utilize spectral analysis in the frequency domain or seasonal pattern detection~\cite{woo2022cost}. 
Specifically, the Fourier Transform decomposes the time series into distinct frequency components, effectively disentangling overlapping periodic information. This decomposition enhances interpretability, making the analysis more straightforward. 
Hence, in this section, we introduce Time-Frequency Synergy Learning to fully harness the complementary strengths of these time and frequency domains. 
It is worth noting that this module is solely designed for the seasonal component because the frequency domain excels at highlighting periodicity, which aligns closely with the nature of seasonal patterns.

To be specific, we first segment the seasonal component $\mathcal{S}\in\mathbb{R}^{N\times \mathcal{L}\times d}$ into sub-level patches $\mathbf{P}_{1:\mathcal{P}}=\{\mathbf{P}_{1},\mathbf{P}_{2},\cdots,\mathbf{P}_\mathcal{P}\}\in\mathbb{R}^{N\times \mathcal{P}\times d}$ with patch length $P$ and stride $S$, where
$\mathcal{P}=\lfloor\frac{(L-P)}{S}\rfloor+2$ denotes the length of patches.
Then,  Discrete Fourier Transform (DFT) is conducted over the patched series to obtain corresponding frequency signals $\mathbf{F}_{1:\mathcal{K}}=\{\mathbf{F}_1, \mathbf{F}_2,\cdots, \mathbf{F}_\mathcal{K}\}\in\mathbb{C}^{N\times \mathcal{K}\times d}$, where $\mathcal{K}=\lfloor\frac{\mathcal{P}}{2}\rfloor+1$, the $\mathbf{F}_k$ can be formulated as:
\begin{equation}
    \mathbf{F}_k=\sum^\mathcal{P}_{p=1}\mathbf{P}_p\cdot e^{-j\frac{2\pi}{\mathcal{P}}kp}=\mathbf{A}_k\cdot e^{j\mathbf{\varphi}_k},
\end{equation}
where $\mathbf{A}_k$ and $\mathbf{\varphi}_k$ denote the corresponding amplitude and phase, respectively.
$j$ is an imaginary unit.

In order to leverage the complementarity of time and frequency domains, we further perform the Inverse Fourier Transform to obtain the time-frequency two-dimensional relationship $\mathbf{TF} \in \mathbb{R}^{N \times \mathcal{K} \times \mathcal{P} \times d}$. Here, the corresponding element $\mathbf{TF}_{p,k}$ for patch $p$ at frequency $k$ can be formulated as:
\begin{equation}
    \mathbf{TF}_{p,k}=\mathbf{A}_k\cdot e^{j(\mathbf{\varphi}_k+\frac{2\pi}{\mathcal{P}}kp)}. 
\end{equation}
Then, \textit{TFKAN} are utilized to learn the time-frequency complementary dependencies, which contain a batch of one-layer KAN.
Each one-layer $\text{KAN}_p$ is utilized for each patch $p$ as follows: 
\begin{equation}
    \mathbf{h}_{tf}^p=\textit{TFKAN}(\mathbf{TF}_{p,k})=\frac{1}{\mathcal{K}}\sum^{\mathcal{K}}_{k=1}\text{KAN}_p(\mathbf{TF}_{p,k}), 
\end{equation}
where $\mathbf{h}_{tf}^p$ is the time-frequency complementary representation of patch $p$. 
In this way, we obtain the representation of all time patches $\mathbf{h}_{tf}=\{\mathbf{h}_{tf}^1,\mathbf{h}_{tf}^2, \cdots, \mathbf{h}_{tf}^{\mathcal{P}}\} \in\mathbb{R}^{N\times \mathcal{P}\times d}$. 
Subsequently, we perform the Unpatch operation to map the series back to its original input length $\mathcal{L}$, resulting in $\mathbf{h}_{TF} \in \mathbb{R}^{N \times \mathcal{L} \times d}$.

\section{Interpretability Analysis}
\label{sec:model_interpre}
Typically, KAN allows the calculation of each node to be intuitively interpreted by converting learnable univariate activation functions on edges into symbolic formulas and summation operations on nodes. Based on the interpretable architecture of KAN, our proposed iTFKAN model realizes the interpretability of prediction results through \textit{sparsification}, \textit{Pruning}, and \textit{Symbolification} operations. Detailed case studies are provided in Section~\ref{Case Studies}.

\paragraph{Sparsification.} 
During the training progress, a sparsification penalty is imposed on all KANs in our proposed iTFKAN. Specifically, for a KAN network, its regularization loss $\ell_{\mathrm{reg}}$ is the sum over the L2 norm of all activation functions in the network: 
\begin{align}
    \ell_{\mathrm{reg}} =\sum_{d=1}^{D}\sum_{j=1}^{J}\sum_{i=1}^{I}||\phi^d_{i,j}||_2,
    ~ ||\phi||_2  =\frac{1}{O}\sum_{o=1}^{O}a_o^2,
\end{align}
where $||\phi||_2$ denote the L2 norm of a univariate activation function $\phi$.
Thus, the loss function of iTFKAN is calculated as follows:
\begin{equation}
\ell_{\mathrm{total}}=\ell_{\mathrm{pred}}+\lambda\left(\sum_{\text{kan}\in \text{KANs}}\ell^\text{kan}_{\mathrm{reg}}\right),
\end{equation}
where $\ell_{\mathrm{pred}}$ is the predicted loss of the overall architecture.
Its calculation depends on different metrics, specifically, the mean square error (MSE) and the symmetric mean absolute percentage error (SMAPE) are utilized for long-term and short-term forecasting, respectively.
$\lambda$ is a hyper-parameter serving to tune the penalty assigned to sparsification.
\paragraph{Pruning.} After sparsification, we prune the full trained KAN networks into smaller sub-networks to highlight the most important parts.
Considering that each time data point of input $\mathbf{X}_t$ plays an important role in time-series prediction, we prune on the edges rather than the network nodes.
After training, the activation function with a high L2 norm plays a significant role in the KAN network.
Therefore, in line with the sparsification criterion, a univariate activation function $\phi$ will be pruned if its L2 norm $||\phi||_2$ is less than a threshold $\tau$.
\paragraph{Symbolification.}
Finally, the univariate activation functions in iTFKAN can be replaced by the existing symbolic formulas to represent the trained model symbolically. 
Specifically, for each activation function $\phi$, its similarity to a symbolic formula is quantified using the coefficient of determination $R^2$.
The symbolic formula $f_s(\cdot)$ with the highest $R^2$ value is selected.
Besides, to address the data shifts and scalings in neural networks, the final symbolization result is presented as $y\approx cf_s(ax+b)+d$ with fitting parameters $[a,b,c,d]$ by linear regression.

\section{Experiment}
\label{sec:Experiments}
To demonstrate the effectiveness of our model, we conducted extensive experiments
on several widely-used time series forecasting benchmarks varying in
scale and domain to test different benchmark methods.

\subsection{Benchmarks and Experimental Settings}
\paragraph{Benchmarks.} 
For long-term time series forecasting, there are 8 well-established benchmarks for comprehensive comparison: ETT (including ETTh1, ETTh2, ETTm1, and ETTm2), Electricity, Traffic, Weather, and Exchange.
For short-term time series forecasting, M4 benchmark~\cite{m4} provides a comprehensive comparison across various time intervals, including yearly, quarterly, monthly, weekly, daily, and hourly.
More detailed information about benchmarks is presented in Appendix~\ref{appendix:datasets}.

\paragraph{Baselines.} Several well-acknowledged and advanced deep time series forecasting models with different backbones are selected as baselines, including (1) MLP-based models: Dlinear~\cite{dlinear}, FreTS~\cite{frets}, and TimeMixer~\cite{timemixer}; (2) Transformer-based models: iTransformer~\cite{itransformer}, PatchTST~\cite{patchtst}, Crossformer~\cite{crossformer}, and Autoformer~\cite{autoformer}); and (3) CNN-based model: TimesNET~\cite{timesnet}.

\paragraph{Experimental Settings.}
To ensure a fair and comprehensive comparison, we closely follow the experimental setup of well-integrated Time Series Library (TSLib)~\cite{wang2024deep}. The parameters used for all baselines are the published optimal settings, and we carefully verify that the results match what they reported. More detailed information about experimental settings is presented in Appendix~\ref{appendix:experimental_setting}. The hyperparameters of iTFKAN are determined by grid search, with the final settings detailed in Appendix~\ref{appendix:hyper_param}.

\subsection{Forecasting Results}
Table~\ref{tab:long-term} shows the overall long-term and short-term time series forecasting experimental results of iTFKAN compared to several outstanding baselines across various widely used datasets. The best in \textcolor{red}{\textbf{bold}} and the second \textcolor{blue}{\underline{underlined}}. We can see that iTFKAN achieves either the top or second-best predictions across all long-term and short-term forecasting datasets. In the Electricity and Traffic datasets, iTFKAN is slightly inferior to TimeMixer and iTransformer. This is due to these two datasets involving hundreds of variables with strong dependencies among them, where intricately designed channel-dependent models with complex learning structures have advantages. Meanwhile, our model maintains channel independence to learn the evolution patterns of univariate series while keeping model complexity relatively low.

\begin{table}[t!]
  \caption{Forecasting Results. We fix the lookback length $L=96$ and prediction lengths $S\in\{96, 192, 336, 720\}$ for long-term forecasting, $S\in\{6, 8, 13, 14,48\}$ for short-term forecasting. Results are averaged from all prediction lengths. A lower MSE, MAE, sMAPE, MASE or OWA indicates a better prediction. See Appendix~\ref{appendix:fullresults} for the full results.}
  \resizebox{\linewidth}{!}{
  \label{tab:long-term}
\begin{tabular}{ccccccccccc}
\toprule
\multirow{3}{*}{Models}      & \multirow{3}{*}{Metrics} & \multirow{2}{*}{\textbf{iTFKAN}} & \multicolumn{3}{c}{MLP-based}& \multicolumn{4}{c}{Transformer-based} & CNN-based\\
\cmidrule(lr){4-6} 
\cmidrule(lr){7-10}
\cmidrule(lr){11-11}
      &   &  \multirow{2}{*}{\textbf{(Ours)}} & TimeMixer & FreTS & Dlinear & iTransformer & PatchTST & Crossformer & Autoformer & TimesNET \\
      &  &  & (2024) & (2024) & (2023) & (2024) & (2023) & (2023) & (2021) & (2023) \\
    \midrule
    & & \multicolumn{8}{c}{Long-term Time Series Forecasting}\\
    \midrule
\multirow{2}{*}{ETTh1}      & MSE     & \textcolor{red}{\textbf{0.434}}& 0.448     & 0.491 & 0.461   & 0.458        & \textcolor{blue}{\underline{0.446}}       & 0.590       & 0.494      & 0.461    \\
   & MAE     & \textcolor{blue}{\underline{0.440}}   & \textcolor{red}{\textbf{0.438}}     & 0.478 & 0.457   & 0.450        & 0.445    & 0.553       & 0.487      & 0.455    \\
    \midrule
\multirow{2}{*}{ETTh2}       & MSE     & \textcolor{red}{\textbf{0.360}}& \textcolor{blue}{\underline{0.372} }       & 0.532 & 0.500   & 0.382        & 0.379    & 0.802       & 0.419      & 0.400    \\
   & MAE     & \textcolor{red}{\textbf{0.394}}& \textcolor{blue}{\underline{0.400}}        & 0.506 & 0.482   & 0.406        & 0.406    & 0.632       & 0.442      & 0.419    \\
    \midrule
\multirow{2}{*}{ETTm1}       & MSE     & \textcolor{red}{\textbf{0.380}}& \textcolor{blue}{\underline{0.384} }       & 0.418 & 0.404   & 0.408        & 0.386    & 0.472       & 0.514      & 0.401    \\
   & MAE     & \textcolor{red}{\textbf{0.395}}& \textcolor{blue}{\underline{0.399}}        & 0.424 & 0.408   & 0.412        & 0.400    & 0.461       & 0.492      & 0.412    \\
    \midrule
\multirow{2}{*}{ETTm2}      & MSE     & \textcolor{red}{\textbf{0.274}}& \textcolor{blue}{\underline{0.276} }       & 0.346 & 0.324   & 0.291        & 0.290    & 0.661       & 0.307      & 0.295    \\
   & MAE     & \textcolor{blue}{\underline{0.324} }  & \textcolor{red}{\textbf{0.323}}     & 0.393 & 0.377   & 0.335        & 0.334    & 0.534       & 0.350      & 0.332    \\
    \midrule
\multirow{2}{*}{Electricity} & MSE     & \textcolor{blue}{\underline{0.185} }  & \textcolor{blue}{\underline{0.185} }       & 0.208 & 0.208   & \textcolor{red}{\textbf{0.175}}        & 0.206    & 0.191       & 0.349      & 0.196    \\
   & MAE     & \textcolor{blue}{\underline{0.273}}   & 0.274     & 0.296 & 0.295   & \textcolor{red}{\textbf{0.267}}        & 0.292    & 0.287       & 0.368      & 0.296    \\
    \midrule
\multirow{2}{*}{Traffic}     & MSE     & 0.503& \textcolor{blue}{\underline{0.481} }       & 0.574 & 0.624   & \textcolor{red}{\textbf{0.422}}        & 0.490    & 0.546       & 0.677      & 0.628    \\
   & MAE     & 0.311& \textcolor{blue}{\underline{0.298}  }      & 0.340 & 0.384   & \textcolor{red}{\textbf{0.283}}        & 0.314    & 0.276       & 0.428      & 0.333    \\
    \midrule
\multirow{2}{*}{Weather}    & MSE     & \textcolor{red}{\textbf{0.244}}& \textcolor{red}{\textbf{0.244}}     & 0.264 & 0.266   & 0.260        & \textcolor{blue}{\underline{0.257} }      & 0.263       & 0.398      & 0.263    \\
   & MAE     & \textcolor{red}{\textbf{0.272}}& \textcolor{blue}{\underline{0.275}}        & 0.311 & 0.317   & 0.280        & 0.281    & 0.322       & 0.422      & 0.288    \\
    \midrule
\multirow{2}{*}{Exchange}   & MSE     & \textcolor{blue}{\underline{0.355} }  & 0.369     & 0.439 & \textcolor{red}{\textbf{0.341}}   & 0.368        & 0.371    & 0.822       & 0.502      & 0.460    \\
   & MAE     & \textcolor{red}{\textbf{0.400}}& \textcolor{blue}{\underline{0.407} }       & 0.457 & 0.415   & 0.408        & 0.410    & 0.655       & 0.500      & 0.469    \\
    \midrule
    & & \multicolumn{8}{c}{Short-term Time Series Forecasting}\\
    \midrule
    \multirow{3}{*}{M4-Yearly}         & sMAPE   & \textcolor{red}{\textbf{13.34}}& \textcolor{blue}{\underline{13.35}}        & 13.58 & 14.36   & 13.71        & 13.54    & 79.33       & 17.47      & 13.49    \\
   & MASE    & \textcolor{red}{\textbf{3.00}} & \textcolor{blue}{\underline{3.03}}& 3.08  & 3.13    & 3.12& 3.05     & 18.69       & 3.94       & 3.07     \\
   & OWA     & \textcolor{red}{\textbf{0.79}} & \textcolor{red}{\textbf{0.79}}& \textcolor{blue}{\underline{0.80}}  & 0.83    & 0.81& \textcolor{blue}{\underline{0.80}}    & 4.78        & 1.03       & \textcolor{red}{\textbf{0.79}}        \\
    \midrule
    \multirow{3}{*}{M4-Quartly}         & sMAPE   & \textcolor{red}{\textbf{10.04}}& 10.19     & 10.29 & 10.50   & 10.48        & 10.78    & 74.42       & 14.88      & \textcolor{blue}{\underline{10.08}}       \\
   & MASE    & \textcolor{red}{\textbf{1.17}} & 1.20      & 1.20  & 1.24    & 1.24& 1.28     & 13.19       & 1.95       & \textcolor{blue}{\underline{1.18}}        \\
   & OWA     & \textcolor{red}{\textbf{0.88}} & 0.90      & 0.91  & 0.93    & 0.93& 0.96     & 8.19        & 1.39       & \textcolor{blue}{\underline{0.89}}        \\
    \midrule
    \multirow{3}{*}{M4-Monthly}         & sMAPE   & \textcolor{red}{\textbf{12.65}}& \textcolor{blue}{\underline{12.76}}        & 13.08 & 13.40   & 13.22        & 14.23    & 68.77       & 18.18      & 12.84    \\
   & MASE    & \textcolor{red}{\textbf{0.92}} & \textcolor{blue}{\underline{0.94}}& 0.99  & 1.01    & 1.02& 1.13     & 11.28       & 1.58       & 0.95     \\
   & OWA     & \textcolor{red}{\textbf{0.87}} & \textcolor{blue}{\underline{0.88}}& 0.92  & 0.94    & 0.94& 1.02     & 7.69        & 1.38       & 0.89     \\
    \midrule
    \multirow{3}{*}{M4-Others}         & sMAPE   & \textcolor{red}{\textbf{4.77}} & 5.05      & 5.41  & 5.11    & 5.15& 5.08     & 176.07      & 6.85       & \textcolor{blue}{\underline{5.03}}        \\
   & MASE    & \textcolor{red}{\textbf{3.16}} & 3.40      & 3.64  & 3.65    & 3.48& 3.30     & 116.65      & 4.93       & \textcolor{blue}{\underline{3.36}}        \\
   & OWA     & \textcolor{red}{\textbf{1.00}} & 1.07      & 1.14  & 1.13    & 1.09& \textcolor{blue}{\underline{1.06}}     & 44.07       & 1.50       & \textcolor{blue}{\underline{1.06}}  \\
\bottomrule
\end{tabular}
}
\end{table}

\subsection{Interpretability Case Studies}
\label{Case Studies}

As stated in Section~\ref{sec:model_interpre}, iTFKAN performs a symbolic transformation based on the KAN backbone, which facilitates straightforward analysis of complex patterns in time series data and provides interpretable insights into the prediction results. Thus, in this section, we investigate the interpretability of iTFKAN through two case studies: \textit{Module Importance Analysis} and \textit{Interpretability Analysis of SeasonalKAN}.

\paragraph{Module Importance Analysis.}

In iTFKAN, activation functions with high L2 norm values are considered important and are preserved during the pruning process. Therefore, the number of preserved edges and the pruning ratio can serve as indicators of the importance of a KAN-based module. Based on this insight, we have listed the preserved edges and pruning ratios of iTFKAN trained on the ETTh1 dataset. As illustrated in Table~\ref{tab:prune_status}, we can see that \textit{SeasonalKAN} and \textit{TFKAN} have more preserved edges compared to \textit{TrendKAN}, underscoring their significance in seasonal information mining. The high number of preserved edges and the pruning ratio demonstrate the efficiency of \textit{TFKAN}. This observation is consistent with the characteristics of the ETTh1 dataset, which exhibits significant periodicity combined with some trends and fluctuations. Furthermore, the high number of preserved edges and low pruning ratio underlines the effectiveness of \textit{TFKAN}, which, by taking patched seasonal information as input, is endowed with greater informativeness.

\paragraph{Interpretability of \textit{SeasonalKAN}.}

\begin{figure}[t!]
	\centering
	\includegraphics[width=\linewidth]{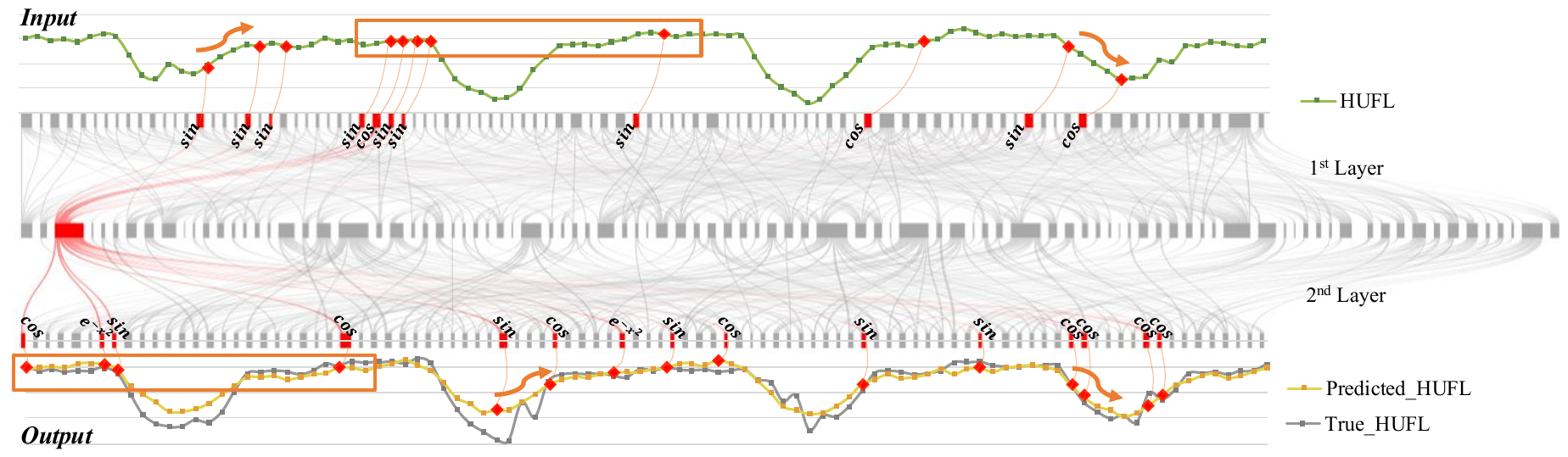}
	\caption{
    The symbolic results and data flow for the \textit{SeasonalKAN} module of the iTFKAN model trained on ETTh1 dataset.
    `HUFL' denotes the variable selected for analysis in ETTh1.
    The red markers label the important inputs and outputs associated with the specified node in layer one.
    The width of the links is proportional to the L2 norm $||\phi||_2$ of the edge, displaying the importance of the edges.
    }
	\label{fig:interpre_etth1}
\end{figure}

Based on the observations of the importance of the \textit{SeasonalKAN} module mentioned above, we further use \textit{SeasonalKAN} as a case study to investigate whether the nodes within this module can capture important temporal patterns and be utilized for downstream forecasting. Specifically, we explore the correlation between nodes with high activation function values and specific patterns in the output predictions.
Figure~\ref{fig:interpre_etth1} displays the visualization results of \textit{SeasonalKAN} trained on the ETTh1 dataset. The full symbolization results of related activation functions are detailed in Appendix~\ref{appendix:symbolic_results}.

Technically, we set the threshold $\tau=5e-4$ and retain the top three significant edges for a visualization based on the L2 norm $||\phi||_2$ of the univariate activation function.
These edges emphasize the data flow between different layers. We highlight an important node in the first layer and its connecting nodes in subsequent layers for further analysis.

As shown in Figure~\ref{fig:interpre_etth1}, the marked input time data points primarily indicate periodic upward, steady, and downward trends, along with periodic intervals. These patterns are significantly reflected in the output data points that share the same layer-one node. This indicates that the specified layer-one node is capable of discovering complex patterns in the time series data and applying them to the forecasting process. Most of the activation functions connected to this layer-one node are periodic, which is consistent with the learning tendency of the seasonality block. Therefore, the detailed contribution of each node can be intuitively displayed through symbolic analysis, which allows researchers to gain insights and further improve the forecasting process.

\begin{table}[t!]
\centering
  \caption{Pruning status of TaylorKANs within iTFKAN on ETTh1 dataset.
  A high threshold $\tau=5e-4$ is set here to facilitate the observation.
  The fixed edge number in \textit{TrendInject} and \textit{PeriodInject} has been subtracted.
  }
  \label{tab:prune_status}
  \begin{tabular}{cccccc}
    \toprule
    KAN & Layer& \#Prune & \#Preserved & \#Total & Prune Ratio \\
    \midrule
    {\textit{TrendKAN}} & 0 & 8,481 & 447 & 8,928& 94.99\%\\
    {\textit{TrendKAN}} & 1 & 9,100 & 116 & 9,216& 98.74\%\\
    {\textit{SeasonalKAN}} & 0 & 7,106 & 1,630 & 8,736& 81.34\%\\
    {\textit{SeasonalKAN}} & 1 & 8,992 & 224 & 9,216& 97.57\%\\
    {\textit{TFKAN}} & - & 90 & 1,287 & 1,377& 6.54\%\\
  \bottomrule
\end{tabular}
\end{table}

\subsection{\textbf{Ablation Study}}

To validate the effectiveness of each component in iTFKAN, we conduct ablation experiments on the ETTm1 dataset and the M4-Monthly dataset for long-term and short-term time series forecasting tasks. We perform removal (\emph{w/o}) or replacement (\emph{Repl.}) operations on each key component. Specifically, \textit{w/o TrendInject} and \textit{w/o SeasonalInject} refer to the removal of the prior knowledge injection strategies in TrendKAN and SeasonalKAN, respectively, while \textit{w/o TFKAN} denotes the removal of the \textit{TFKAN} module. \textit{Repl. TFKAN as FreqKAN} refers to replacing the TFKAN module with FreqKAN, which leverages the approach proposed in FreTS~\cite{frets}. Unlike TFKAN, which captures time-frequency complementarities, \textit{Repl. TFKAN as FreqKAN} directly combine the information from time and frequency domains.

\begin{figure}[t!]
	\centering
	\includegraphics[width=\linewidth]{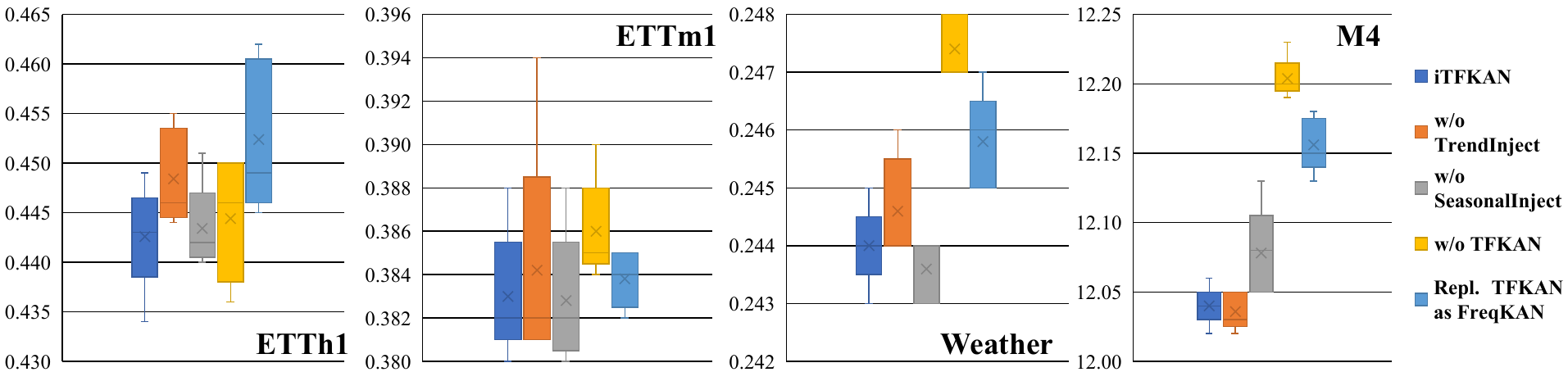}
	\caption{
    The Experimental Results of iTFKAN, its ablation variants, and TimeMixer on ETTh1, ETTm1, Weather and M4 datasets, all conducted with 5 seeds. Average scores are marked as ‘×’.
    }
	\label{fig:ablabtion}
\end{figure}

As shown in Figure~\ref{fig:ablabtion}, we can make the following observations: 
\begin{itemize}

\item \textbf{Performance and Robustness of iTFKAN}. iTFKAN achieves top average performance across nearly all datasets, enhanced by injected knowledge that strengthens its robustness. The ablation of any component will, to varying degrees, impact its performance or robustness. Removing any component will impact its performance or robustness to varying degrees, highlighting the rationality of model design.
\item \textbf{Prior Knowledge Injection (\textit{TrendInject} \& \textit{SeasonalInject})}. Although less effective on highly periodic ETTh1, ETTm1, and Weather, \textit{SeasonalInject} excels in M4, where short-term forecasting struggles with periodicity.
While \textit{TrendInject} compensates for KAN's neglect of trend information in highly periodic datasets, this observation effectively validates that our prior knowledge injection into iTFKAN can significantly alleviate KAN's data-driven nature.
\item \textbf{Time-Frequency Synergy Learning (\textit{TFKAN})}. \textit{TFKAN} plays a critical role in iTFKAN. Removing \textit{iTFKAN} leads to substantial performance declines, and in some cases (e.g., the ETTh1 dataset), replacing it with \textit{FreqKAN} results in even worse performance. This suggests that directly integrating time-domain and frequency-domain information may lead to information loss. In contrast, the Time-Frequency Synergy Learning enables a smoother integration by leveraging the two-dimensional time-frequency dependencies.
\end{itemize}

\section{Conclusion}
\label{sec:conclusion}

In this paper, we propose a novel interpretable framework, iTFKAN, for time series forecasting. We introduce two key strategies: prior knowledge injection and time-frequency synergy learning, to guide the learning process of KAN for time series forecasting. 
iTFKAN achieves both outstanding predictive performance and exceptional interpretability. Extensive experiments demonstrate its superiority over state-of-the-art baselines. Additionally, iTFKAN provides interpretability through symbolic transformation, offering an intuitive analysis of the model's decision-making rationale. This framework also lays a foundation for future research on interpretability in time series prediction.

\bibliographystyle{unsrt}
\small
\bibliography{main}
\normalsize

\newpage


\appendix

\begin{table*}[t!]
\centering
  \caption{{Statistics of the overall Datasets used in this paper. \#Subsets is the number of the subsets, \#Variate denotes the number of variates of dataset, and $L$ (Predict) denotes the prediction length in the forecasting task.}
  }
  \label{tab:data_char}
  
\scalebox{1}{
  \begin{tabular}{ccccccc}
    \toprule
    Tasks& Dataset & \#Subsets & Domain & \#Variate & $L$ (Predict) & Frequency \\
    \midrule
    \multirow{5}{*}{Long-term} & ETT & 4 & Electricity& 7 & 96\textasciitilde720 & 15min\\
     & Electricity & - & Electricity& 321 & 96\textasciitilde 720 & Hourly\\
     & Traffic & - & Transportation & 862 & 96\textasciitilde 720 & Hourly\\
     & Weather & - & Weather& 21 & 96\textasciitilde 720 & 10min\\
     & Exchange & - & Exchange Rates & 8 & 96\textasciitilde 720 & Daily\\
    \midrule
    {Short-term}  & M4 & 6 & Database & 1 & 6\textasciitilde 48 & Hourly\textasciitilde Yearly\\
  \bottomrule
\end{tabular}
}
\end{table*}

\section{Experimental Setting}
\label{appendix:experimental_setting}
\subsection{Datasets}
\label{appendix:datasets}
The statistics presented in Table~\ref{tab:data_char} encompass the overall datasets utilized in this paper, which include both long-term and short-term datasets for time series prediction.

\subsection{\textbf{Implementation Details}}

All experiments are conducted on NVIDIA GeForce RTX 3090 24GB GPU and NVIDIA GeForce RTX A6000 48GB GPU. 
And for long-term time series forecasting task, we present MAE (Mean Absolute Errors) and MSE (Mean Squared Errors) results as the evaluation metrics.
While for short-term time series forecasting task, we present sMAPE (symmetric Mean Absolute Percentage Error), MASE (Mean Absolute Scaled Error) and OWA (Overall Weighted Average Error) results as the the evaluation metrics.

\subsection{\textbf{Hyper-parameters of iTFKAN}}
\label{appendix:hyper_param}
We fix the $K=5$ in \textit{SeasonalInject} and $\phi=0.01$ in the total loss calculation.
For the other hyper-parameters of iTFKAN, we use grid search to determine the best value, the result is shown in Table~\ref{hyper-param}.

\begin{table}[h]
\centering
\caption{\label{hyper-param}
Grid search results of all hyper-parameters in iTFKAN on different datasets.
$d$ denotes the expansion dimension of $\bf{X}$ to $\mathcal{X}$.
$\eta$ is the learning rate of iTFKAN and 
$|B|$ is the batch size of iTFKAN.
$S$ and $P$ denote the stride and patch length in the patch operation in the time-frequency synergy learning.
}
\scalebox{1}{
\begin{tabular}{cccccc}
\toprule
Hyper-params & $d$ & $|B|$  & $\eta$      & $S$ & $P$ \\
\midrule
ETTh1        & 32         & 64 & 0.0005  & 6      & 6        \\
ETTh2        & 64         & 16 & 0.00005 & 4      & 6        \\
ETTm1        & 16         & 16 & 0.0001  & 12     & 4        \\
ETTm2        & 64         & 16 & 0.00005 & 6      & 8        \\
Electricity  & 32         & 8  & 0.001   & 8      & 6        \\
Traffic      & 32         & 4  & 0.0005  & 6      & 6        \\
Weather      & 16         & 64 & 0.001   & 8      & 6        \\
Exchange     & 32         & 32 & 0.0005  & 6      & 6        \\
M4           & 32         & 8  & 0.001   & 2      & 4      \\
\bottomrule
\end{tabular}}

\end{table}

\section{Interpretability Cases}
\subsection{Symbolic Results of Seasonality Block}
\label{appendix:symbolic_results}
Table~\ref{tab:symbolic_results} shows the full symbolic formula results of \textit{SeasonalKAN}.

\begin{table}[t!]
\centering
  \caption{The full symbolic formula results of \textit{SeasonalKAN}. $i$ and $j$ respectively represent the indices on the input dimension and the output dimension, specifying the corresponding activation function edge.
  }
  \scalebox{1}{
  \label{tab:symbolic_results}
  \begin{tabular}{cccc}
    \toprule
    Layer& $i$ & $j$ & Symbolic Formula \\
    \midrule
    1st & 14 & 2& $114.33sin(-0.02x-4.70)-114.31$ \\
    1st & 18 & 2& $-0.76sin(-0.12x-7.40)-0.67$\\
    1st & 20 & 2& $35.64sin(-0.03x-1.61)+35.63$\\
    1st & 28 & 2& $-49.59sin(-0.03x+1.61)+49.58$\\
    1st & 29 & 2& $-82.09cos(-0.03x+0.03)+82.06$\\
    1st & 30 & 2& $4.34sin(-0.09x-1.40)+4.29$\\
    1st & 31 & 2& $21.10sin(-0.04x-7.80)+21.08$\\
    1st & 14 & 2& $-11.01sin(-0.05x-7.79)-10.98$ \\
    1st & 14 & 2& $32.25cos(-0.03x-9.39)+32.25$ \\
    1st & 14 & 2& $-2.58sin(-0.09x-1.40)-2.52$ \\
    1st & 14 & 2& $3.60cos(-0.08x-3.00)+3.57$ \\
    \midrule
    2nd & 2 & 0& $0.94cos(-0.11x-6.00)-0.91$ \\
    2nd & 2 & 6& $-0.79e^{-{(-1.34x-1.34)}^2}+0.13$  \\
    2nd & 2 & 7&  $-33.70sin(-0.04x-1.58)-33.70$\\
    2nd & 2 & 24& $26.64cos(-0.04x-3.11)+26.63$ \\
    2nd & 2 & 36& $-28.86sin(-0.04x-4.69)+28.86$ \\
    2nd & 2 & 40& $10.09cos(-0.05x-9.37)+10.07$\\
    2nd & 2 & 45& $-2.31e^{-{(-0.03x+1.18)}^2}+0.59$  \\
    2nd & 2 & 49& $-18.97sin(-0.04x-4.69)+18.97$ \\
    2nd & 2 & 53&  $34.15cos(-0.04x)-34.14$\\
    2nd & 2 & 64&  $-23.38sin(-0.04x+4.73)-23.38$\\
    2nd & 2 & 73&  $-4.45sin(-0.05x-7.78)-4.46$\\
    2nd & 2 & 80&  $17.04cos(-0.04x-3.11)+17.04$\\
    2nd & 2 & 81&  $22.38cos(-0.06x-9.42)+22.37$\\
    2nd & 2 & 86&  $27.95cos(-0.04x-9.46)+27.93$\\
    2nd & 2 & 87&  $18.94cos(-0.04x+6.31)-18.93$\\
  \bottomrule
\end{tabular}
}
\end{table}

\section{Full Results}
\label{appendix:fullresults}

In this section, we provide the full results of both long-term and short-term forecasting experiments in Table~\ref{tab:long-term-all} and Table~\ref{tab:short-term}, respectively.

\begin{table*}
  \caption{All Results about Experiments on Long-term Time Series Forecasting. The best in \textcolor{red}{\textbf{bold}} and the second \textcolor{blue}{\underline{underlined}}.}
  \resizebox{\linewidth}{!}{
  \label{tab:long-term-all}
\begin{tabular}{cccccccccccccccccccc}
    \toprule
Models      &     & \multicolumn{2}{c}{\textbf{iTFKAN}}      & \multicolumn{2}{c}{TimeMixer}   & \multicolumn{2}{c}{FreTS} & \multicolumn{2}{c}{Dlinear}     & \multicolumn{2}{c}{iTransformer} & \multicolumn{2}{c}{PatchTST}    & \multicolumn{2}{c}{Crossformer} & \multicolumn{2}{c}{Autoformer} & \multicolumn{2}{c}{TimesNET} \\
    \midrule
Metrics     &     & MSE & MAE & MSE & MAE & MSE & MAE      & MSE & MAE & MSE  & MAE & MSE & MAE & MSE    & MAE & MSE & MAE& MSE& MAE  \\
    \midrule
\multirow{5}{*}{ETTh1}       & 96  & \textcolor{red}{\textbf{0.385}} & \textcolor{blue}{\underline{0.404} }   & \textcolor{blue}{\underline{0.386} }   & \textcolor{red}{\textbf{0.400}} & 0.402  & 0.416    & 0.397  & 0.412  & 0.386& 0.405  & 0.384  & \textcolor{blue}{\underline{0.404} }   & 0.422  & 0.443       & 0.464  & 0.469 & 0.389 & 0.412\\
 & 192 & 0.434  & 0.436  & \textcolor{red}{\textbf{0.428}} & \textcolor{red}{\textbf{0.426}} & 0.472  & 0.462    & 0.446  & 0.441  & 0.440& 0.436  & 0.434  & 0.437  & 0.492  & 0.485       & 0.508  & 0.488 & 0.440 & 0.442\\
 & 336 & \textcolor{red}{\textbf{0.466}} & \textcolor{red}{\textbf{0.451}} & 0.489  & 0.454  & 0.518  & 0.484    & \textcolor{blue}{\underline{0.488}}    & 0.466  & 0.490& 0.460  & \textcolor{red}{\textbf{0.466}} & \textcolor{blue}{\underline{0.452}}    & 0.632  & 0.584       & 0.495  & 0.486 & 0.495 & 0.470\\
 & 720 & \textcolor{red}{\textbf{0.451}} & \textcolor{red}{\textbf{0.467}} & \textcolor{blue}{\underline{0.487} }   & \textcolor{blue}{\underline{0.472} }   & 0.573  & 0.548    & 0.511  & 0.510  & 0.516& 0.497  & 0.501  & 0.486  & 0.815  & 0.698       & 0.508  & 0.506 & 0.520 & 0.495\\
 & avg & \textcolor{red}{\textbf{0.434}} & \textcolor{blue}{\underline{0.440} }   & 0.448  & \textcolor{red}{\textbf{0.438}} & 0.491  & 0.478    & 0.461  & 0.457  & 0.458& 0.450  & \textcolor{blue}{\underline{0.446} }   & 0.445  & 0.590  & 0.553       & 0.494  & 0.487 & 0.461 & 0.455\\
    \midrule
\multirow{5}{*}{ETTh2}       & 96  & \textcolor{red}{\textbf{0.278}} & \textcolor{red}{\textbf{0.334}} & 0.303  & 0.351  & 0.347  & 0.399    & 0.325  & 0.380  & 0.300& 0.350  & \textcolor{blue}{\underline{0.294} }   & \textcolor{blue}{\underline{0.348} }   & 0.635  & 0.555       & 0.336  & 0.381 & 0.332 & 0.370\\
 & 192 & \textcolor{red}{\textbf{0.350}} & \textcolor{red}{\textbf{0.382}} & \textcolor{blue}{\underline{0.370} }   & \textcolor{blue}{\underline{0.393} }   & 0.480  & 0.478    & 0.457  & 0.459  & 0.381& 0.399  & 0.371  & 0.395  & 0.578  & 0.548       & 0.437  & 0.442 & 0.397 & 0.410\\
 & 336 & \textcolor{red}{\textbf{0.396}} & \textcolor{red}{\textbf{0.420}} & \textcolor{blue}{\underline{0.401}}    & \textcolor{blue}{\underline{0.422}}    & 0.519  & 0.509    & 0.491  & 0.487  & 0.419& 0.430  & 0.418  & 0.430  & 0.896  & 0.669       & 0.453  & 0.471 & 0.446 & 0.448\\
 & 720 & \textcolor{blue}{\underline{0.414} }   & \textcolor{blue}{\underline{0.440}}    & \textcolor{red}{\textbf{0.412}} & \textcolor{red}{\textbf{0.434}} & 0.780  & 0.638    & 0.727  & 0.603  & 0.428& 0.445  & 0.431  & 0.450  & 1.097  & 0.757       & 0.451  & 0.474 & 0.435 & 0.449\\
 & avg & \textcolor{red}{\textbf{0.360}} & \textcolor{red}{\textbf{0.394}} & \textcolor{blue}{\underline{0.372}}    & \textcolor{blue}{\underline{0.400}}    & 0.532  & 0.506    & 0.500  & 0.482  & 0.382& 0.406  & 0.379  & 0.406  & 0.802  & 0.632       & 0.419  & 0.442 & 0.400 & 0.419\\
    \midrule
\multirow{5}{*}{ETTm1}       & 96  & \textcolor{red}{\textbf{0.317}} & \textcolor{red}{\textbf{0.360}} & \textcolor{blue}{\underline{0.321}}    & \textcolor{blue}{\underline{0.361}}    & 0.352  & 0.385    & 0.343  & 0.371  & 0.344& 0.378  & 0.326  & 0.365  & 0.403  & 0.412       & 0.464  & 0.470 & 0.347 & 0.383\\
 & 192 & \textcolor{red}{\textbf{0.357}} & \textcolor{red}{\textbf{0.379}} & \textcolor{blue}{\underline{0.367} }   & \textcolor{blue}{\underline{0.387} }   & 0.394  & 0.406    & 0.382  & 0.390  & 0.381& 0.395  & \textcolor{blue}{\underline{0.367}}    & \textcolor{blue}{\underline{0.387} }   & 0.477  & 0.458       & 0.514  & 0.484 & 0.382 & 0.399\\
 & 336 & \textcolor{red}{\textbf{0.389}} & \textcolor{red}{\textbf{0.401}} & \textcolor{blue}{\underline{0.391}}    & \textcolor{blue}{\underline{0.403}}    & 0.430  & 0.431    & 0.415  & 0.417  & 0.419& 0.419  & 0.398  & 0.407  & 0.474  & 0.472       & 0.522  & 0.495 & 0.409 & 0.420\\
 & 720 & \textcolor{red}{\textbf{0.456}} & \textcolor{red}{\textbf{0.440}} & 0.457  & 0.444  & 0.494  & 0.472    & 0.474  & 0.453  & 0.486& 0.456  & \textcolor{blue}{\underline{0.454}}    & \textcolor{blue}{\underline{0.442}}    & 0.532  & 0.503       & 0.554  & 0.518 & 0.465 & 0.447\\
 & avg & \textcolor{red}{\textbf{0.380}} & \textcolor{red}{\textbf{0.395}} & \textcolor{blue}{\underline{0.384}}    & \textcolor{blue}{\underline{0.399}}    & 0.418  & 0.424    & 0.404  & 0.408  & 0.408& 0.412  & 0.386  & 0.400  & 0.472  & 0.461       & 0.514  & 0.492 & 0.401 & 0.412\\
    \midrule
\multirow{5}{*}{ETTm2}       & 96  & \textcolor{red}{\textbf{0.174}} & \textcolor{red}{\textbf{0.257}} & \textcolor{blue}{\underline{0.176} }   & \textcolor{blue}{\underline{0.258} }   & 0.194  & 0.290    & 0.189  & 0.283  & 0.184& 0.268  & 0.186  & 0.268  & 0.285  & 0.370       & 0.210  & 0.291 & 0.189 & 0.266\\
 & 192 & \textcolor{blue}{\underline{0.238} }   & \textcolor{blue}{\underline{0.301}}    & \textcolor{red}{\textbf{0.236}} & \textcolor{red}{\textbf{0.298}} & 0.283  & 0.359    & 0.266  & 0.338  & 0.253& 0.313  & 0.246  & 0.307  & 0.388  & 0.438       & 0.272  & 0.330 & 0.252 & 0.307\\
 & 336 & \textcolor{blue}{\underline{0.299}}    & \textcolor{blue}{\underline{0.340}}    & \textcolor{red}{\textbf{0.298}} & \textcolor{red}{\textbf{0.338}} & 0.360  & 0.407    & 0.388  & 0.430  & 0.314& 0.351  & 0.307  & 0.346  & 0.613  & 0.541       & 0.323  & 0.361 & 0.321 & 0.349\\
 & 720 & \textcolor{red}{\textbf{0.394}} & \textcolor{red}{\textbf{0.398}} & \textcolor{red}{\textbf{0.394}} & \textcolor{red}{\textbf{0.398}} & 0.545  & 0.516    & 0.454  & 0.456  & \textcolor{blue}{\underline{0.411} }    & \textcolor{blue}{\underline{0.406}}    & 0.420  & 0.413  & 1.356  & 0.785       & 0.423  & 0.419 & 0.419 & \textcolor{blue}{\underline{0.406}}  \\
 & avg & \textcolor{red}{\textbf{0.276}} & \textcolor{blue}{\underline{0.324} }   & \textcolor{red}{\textbf{0.276}} & \textcolor{red}{\textbf{0.323}} & 0.346  & 0.393    & 0.324  & 0.377  & 0.291& 0.335  & \textcolor{blue}{\underline{0.290} }   & 0.334  & 0.661  & 0.534       & 0.307  & 0.350 & 0.295 & 0.332\\
    \midrule
\multirow{5}{*}{Electricity} & 96  & 0.158  & 0.248  & 0.156  & \textcolor{blue}{\underline{0.247} }   & 0.189  & 0.276    & 0.194  & 0.277  & \textcolor{red}{\textbf{0.148}}  & \textcolor{red}{\textbf{0.240}} & 0.185  & 0.272  & \textcolor{blue}{\underline{0.152}}       & 0.252       & 0.213  & 0.328 & 0.171 & 0.274\\
 & 192 & 0.169  & \textcolor{blue}{\underline{0.259}}    & 0.169  & 0.260  & 0.191  & 0.279    & 0.190  & 0.278  & \textcolor{red}{\textbf{0.165}}  & \textcolor{red}{\textbf{0.257}} & 0.187  & 0.273  & \textcolor{blue}{\underline{0.164} }      & 0.264       & 0.248  & 0.348 & 0.180 & 0.284\\
 & 336 & \textcolor{blue}{\underline{0.186} }   & \textcolor{blue}{\underline{0.276}}    & \textcolor{blue}{\underline{0.186}}    & 0.277  & 0.206  & 0.296    & 0.205  & 0.295  & \textcolor{red}{\textbf{0.177}}  & \textcolor{red}{\textbf{0.271}} & 0.205  & 0.294  & \textcolor{blue}{\underline{0.186}}       & 0.285       & 0.238  & 0.350 & 0.200 & 0.300\\
 & 720 & \textcolor{blue}{\underline{0.225} }   & \textcolor{blue}{\underline{0.309}  }  & 0.227  & 0.312  & 0.246  & 0.332    & 0.241  & 0.328  & \textcolor{red}{\textbf{0.210}}  & \textcolor{red}{\textbf{0.300}} & 0.247  & 0.327  & 0.260  & 0.347       & 0.698  & 0.445 & 0.234 & 0.324\\
 & avg & \textcolor{blue}{\underline{0.185} }   & \textcolor{blue}{\underline{0.273} }   & \textcolor{blue}{\underline{0.185} }   & 0.274  & 0.208  & 0.296    & 0.208  & 0.295  & \textcolor{red}{\textbf{0.175}}  & \textcolor{red}{\textbf{0.267}} & 0.206  & 0.292  & 0.191  & 0.287       & 0.349  & 0.368 & 0.196 & 0.296\\
    \midrule
\multirow{5}{*}{Traffic}     & 96  & 0.477  & 0.303  & \textcolor{blue}{\underline{0.459} }   & \textcolor{blue}{\underline{0.287}}    & 0.557  & 0.329    & 0.648  & 0.396  & \textcolor{red}{\textbf{0.393}}  & \textcolor{red}{\textbf{0.270}} & 0.467  & 0.303  & 0.502  & 0.217       & 0.698  & 0.445 & 0.591 & 0.315\\
 & 192 & 0.489  & 0.307  & \textcolor{blue}{\underline{0.463} }   & \textcolor{blue}{\underline{0.287}}    & 0.569  & 0.338    & 0.598  & 0.370  & \textcolor{red}{\textbf{0.411}}  & \textcolor{red}{\textbf{0.277}} & 0.477  & 0.307  & 0.544  & 0.292       & 0.720  & 0.457 & 0.619 & 0.323\\
 & 336 & 0.504  & 0.312  & 0.500  & 0.316  & 0.566  & 0.337    & 0.605  & 0.373  & \textcolor{red}{\textbf{0.424}}  & \textcolor{red}{\textbf{0.283}} & \textcolor{blue}{\underline{0.492}}    & \textcolor{blue}{\underline{0.314}}    & 0.555  & 0.287       & 0.625  & 0.394 & 0.637 & 0.339\\
 & 720 & 0.540  & 0.331  & \textcolor{blue}{\underline{0.500} }   & \textcolor{blue}{\underline{0.303} }   & 0.603  & 0.357    & 0.646  & 0.395  & \textcolor{red}{\textbf{0.459}}  & \textcolor{red}{\textbf{0.301}} & 0.524  & 0.330  & 0.582  & 0.307       & 0.663  & 0.414 & 0.663 & 0.353\\
 & avg & 0.503  & 0.311  & \textcolor{blue}{\underline{0.481} }   & \textcolor{blue}{\underline{0.298} }   & 0.574  & 0.340    & 0.624  & 0.384  & \textcolor{red}{\textbf{0.422}}  & \textcolor{red}{\textbf{0.283}} & 0.490  & 0.314  & 0.546  & 0.276       & 0.677  & 0.428 & 0.628 & 0.333\\
    \midrule
\multirow{5}{*}{Weather}     & 96  & \textcolor{red}{\textbf{0.162}} & \textcolor{red}{\textbf{0.207}} & \textcolor{red}{\textbf{0.162}} & \textcolor{blue}{\underline{0.209} }   & 0.183  & 0.238    & 0.195  & 0.254  & 0.173& 0.214  & 0.174  & 0.218  & \textcolor{blue}{\underline{0.171} }      & 0.237       & 0.275  & 0.345 & 0.173 & 0.220\\
 & 192 & \textcolor{red}{\textbf{0.209}} & \textcolor{red}{\textbf{0.251}} & \textcolor{red}{\textbf{0.209}} & \textcolor{blue}{\underline{0.252} }   & 0.251  & 0.312    & 0.239  & 0.299  & 0.227& 0.258  & 0.221  & 0.257  & 0.230  & 0.300       & 0.326  & 0.374 & \textcolor{blue}{\underline{0.219}  } & 0.260\\
 & 336 & \textcolor{red}{\textbf{0.264}} & \textcolor{red}{\textbf{0.290}} & \textcolor{red}{\textbf{0.264}} & \textcolor{blue}{\underline{0.294}}    & \textcolor{blue}{\underline{0.272} }   & 0.316    & 0.282  & 0.331  & 0.281& 0.299  & 0.281  & 0.301  & 0.278  & 0.339       & 0.561  & 0.531 & 0.285 & 0.306\\
 & 720 & \textcolor{red}{\textbf{0.341}} & \textcolor{red}{\textbf{0.341}} & \textcolor{red}{\textbf{0.341}} & \textcolor{blue}{\underline{0.344} }   & 0.349  & 0.377    & \textcolor{blue}{\underline{0.346}}    & 0.382  & 0.358& 0.350  & 0.353  & 0.349  & 0.371  & 0.410       & 0.430  & 0.439 & 0.376 & 0.365\\
 & avg & \textcolor{red}{\textbf{0.244}} & \textcolor{red}{\textbf{0.272}} & \textcolor{red}{\textbf{0.244}} & \textcolor{blue}{\underline{0.275}}    & 0.264  & 0.311    & 0.266  & 0.317  & 0.260& 0.280  & \textcolor{blue}{\underline{0.257}}    & 0.281  & 0.263  & 0.322       & 0.398  & 0.422 & 0.263 & 0.288\\
    \midrule
\multirow{5}{*}{Exchange}    & 96  & 0.088  & \textcolor{blue}{\underline{0.200}  }  & \textcolor{blue}{\underline{0.082}}    & \textcolor{red}{\textbf{0.199}} & 0.094  & 0.222    & 0.094  & 0.228  & 0.086& 0.206  & \textcolor{red}{\textbf{0.081}} & \textcolor{blue}{\underline{0.200}}    & 0.246  & 0.359       & 0.157  & 0.286 & 0.122 & 0.255\\
 & 192 & \textcolor{red}{\textbf{0.173}} & \textcolor{red}{\textbf{0.295}} & \textcolor{blue}{\underline{0.178} }   & \textcolor{blue}{\underline{0.298} }   & 0.222  & 0.350    & 0.185  & 0.325  & 0.180& 0.303  & 0.208  & 0.326  & 0.465  & 0.502       & 0.283  & 0.392 & 0.256 & 0.361\\
 & 336 & \textcolor{blue}{\underline{0.350} }   & \textcolor{blue}{\underline{0.430} }   & 0.352  & \textcolor{blue}{\underline{0.430}}    & 0.431  & 0.492    & 0.336  & 0.444  & 0.335& 0.419  & \textcolor{red}{\textbf{0.296}} & \textcolor{red}{\textbf{0.396}} & 0.981  & 0.762       & 0.452  & 0.502 & 0.432 & 0.479\\
 & 720 & \textcolor{blue}{\underline{0.811}}    & \textcolor{blue}{\underline{0.676} }   & 0.865  & 0.701  & 1.007  & 0.764    & \textcolor{red}{\textbf{0.749}} & \textcolor{red}{\textbf{0.664}} & 0.869& 0.705  & 0.898  & 0.716  & 1.596  & 0.996       & 1.116  & 0.819 & 1.031 & 0.780\\
 & avg & \textcolor{blue}{\underline{0.355} }   & \textcolor{red}{\textbf{0.400}} & 0.369  & \textcolor{blue}{\underline{0.407}}    & 0.439  & 0.457    & \textcolor{red}{\textbf{0.341}} & 0.415  & 0.368& 0.408  & 0.371  & 0.410  & 0.822  & 0.655       & 0.502  & 0.500 & 0.460 & 0.469       \\
 \bottomrule
\end{tabular}
  }
\end{table*}

\begin{table*}
  \caption{All Results about Experiments on Short-term Time Series Forecasting. The best in \textcolor{red}{\textbf{bold}} and the second \textcolor{blue}{\underline{underlined}}.}
  \resizebox{\linewidth}{!}{
  \label{tab:short-term}
  \begin{tabular}{cccccccccccc}
    \toprule
Models& Metrics & \textbf{iTFKAN} & TimeMixer     & TimesNET      & Autoformer & Crossformer & DLinear & FreTS      & FEDformer  & PatchTST& iTransformer \\ \midrule
\multirow{3}{*}{M4-Yearly}    & sMAPE   & \textcolor{red}{\textbf{13.34}}  & \textcolor{blue}{\underline{13.35} }  & 13.49 & 17.47      & 79.33& 14.36   & 13.58      & 13.67      & 13.54  & 13.71\\
      & MASE    & \textcolor{red}{\textbf{3.00} }  & \textcolor{blue}{\underline{3.03} }   & 3.07  & 3.94& 18.69& 3.13    & 3.08& 3.10& 3.05   & 3.12 \\
      & OWA     & \textcolor{red}{\textbf{0.79}}   & \textcolor{red}{\textbf{0.79}} & \textcolor{red}{\textbf{0.79}} & 1.03& 4.78& 0.83    & \textcolor{blue}{\underline{0.80}} & 0.81& \textcolor{blue}{\underline{0.80}}     & 0.812\\ \midrule
\multirow{3}{*}{M4-Quarterly} & sMAPE   & \textcolor{red}{\textbf{10.04}}  & 10.19 & \textcolor{blue}{\underline{10.08}}   & 14.88      & 74.42& 10.50   & 10.29      & 10.81      & 10.78  & 10.47\\
      & MASE    & \textcolor{red}{\textbf{1.17} }  & 1.20  & \textcolor{blue}{\underline{1.18} }   & 1.95& 13.19& 1.24    & 1.20& 1.30& 1.28   & 1.24 \\
      & OWA     & \textcolor{red}{\textbf{0.88}}   & 0.90  & \textcolor{blue}{\underline{0.89} }   & 1.50& 8.19& 0.93    & 0.91& 0.96& 0.96   & 0.93 \\ \midrule
\multirow{3}{*}{M4-Monthly}   & sMAPE   & \textcolor{red}{\textbf{12.65}}  & \textcolor{blue}{\underline{12.76}}   & 12.84 & 18.18      & 68.77& 13.40   & 13.08      & 14.02      & 14.23  & 12.22\\
      & MASE    & \textcolor{red}{\textbf{0.92}}   & \textcolor{blue}{\underline{0.94} }   & 0.95  & 1.58& 11.28& 1.01    & 0.99& 1.07& 1.13   & 1.02 \\
      & OWA     & \textcolor{red}{\textbf{0.87}}   & \textcolor{blue}{\underline{0.88} }   & 0.89  & 1.38& 7.69& 0.94    & 0.92& 0.99& 1.02   & 0.94 \\ \midrule
\multirow{3}{*}{M4-Weekly}    & sMAPE   & \textcolor{blue}{\underline{10.40} }    & 12.21 & \textcolor{blue}{\underline{10.66} }  & 12.72      & 198.36      & 11.89   & 11.32      & 9.60& \textcolor{red}{\textbf{10.20}} & 10.79\\
      & MASE    & \textcolor{red}{\textbf{2.98} } & 3.77  & 3.20  & 4.93& 98.92& 4.59    & 3.74& 3.38& \textcolor{blue}{\underline{3.03}  }   & 4.00 \\
      & OWA     & \textcolor{red}{\textbf{1.10} }  & 1.35  & 1.16  & 1.58& 28.64& 1.47    & 1.29& 1.13& \textcolor{red}{\textbf{1.10}}  & 1.31 \\ \midrule
\multirow{3}{*}{M4-Daily}     & sMAPE   & \textcolor{red}{\textbf{3.00}}   & 3.11  & 3.18  & 4.36& 178.95      & 3.35    & 3.26& 3.07& \textcolor{blue}{\underline{3.06} }    & 3.24 \\
      & MASE    & \textcolor{red}{\textbf{3.20}}   & 3.35  & 3.44  & 4.84& 125.80      & 3.66    & 3.51& 3.25& \textcolor{blue}{\underline{3.24} }    & 3.45 \\
      & OWA     & \textcolor{red}{\textbf{0.98} }  & 1.02  & 1.05  & 1.45& 48.57& 1.11    & 1.07& \textcolor{blue}{\underline{1.00}} & \textcolor{blue}{\underline{1.00}}     & 1.06 \\ \midrule
\multirow{3}{*}{M4-Hourly}    & sMAPE   & \textcolor{red}{\textbf{17.90}}  & \textcolor{blue}{\underline{18.65}}   & 19.00 & 27.16      & 127.37      & 17.17   & 22.27      & 18.84      & 21.31  & 19.75\\
      & MASE    & \textcolor{blue}{\underline{2.98}}      & 3.59  & \textcolor{red}{\textbf{2.74}} & 5.75& 38.60& 2.78    & 4.83& 2.67& 4.11   & 3.41 \\
      & OWA     & \textcolor{blue}{\underline{1.11} }     & 1.26  & \textcolor{red}{\textbf{1.09}} & 1.94& 11.52& 1.05    & 1.61& 1.07& 1.44   & 1.25 \\ \midrule
\multirow{3}{*}{M4-Avg}& sMAPE   & \textcolor{red}{\textbf{12.02}}  & 12.15 & \textcolor{blue}{\underline{12.19} }  & 17.00      & 77.92& 12.51   & 12.41      & 12.95      & 13.04  & 12.53\\
      & MASE    & \textcolor{red}{\textbf{1.73}}   & \textcolor{blue}{\underline{1.78}}    & \textcolor{blue}{\underline{1.78} }   & 2.63& 18.71& 1.68    & 1.83& 1.86& 1.88   & 1.85 \\
      & OWA     & \textcolor{red}{\textbf{0.85}}   & \textcolor{blue}{\underline{0.87}}    & 0.88  & 1.25& 7.77& 0.90    & 0.89& 0.92& 0.93   & 0.90 \\ 
      \bottomrule
\end{tabular}
  }
\end{table*}


\newpage

\end{document}